\documentclass[sigconf]{acmart}

\usepackage{multirow}
\usepackage{makecell}

\newtheorem{myDef}{Definition}

\newtheorem*{problem}{Problem Statement}

\newcommand{\eg}{\emph{e.g., }}
\newcommand{\ie}{\emph{i.e., }}

\newcommand{\eat}[1]{}
\ifodd 1

\newcommand{\TODO}[1]{{\color{red}TODO:{#1}}}
\newcommand\beftext[1]{{\color[rgb]{0.5,0.5,0.5}{BEFORE:#1}}}
\else

\newcommand{\TODO}[1]{}
\newcommand{\beftext}[1]{}
\fi
\AtBeginDocument{%
  }

\copyrightyear{2025}
\acmYear{2025}
\setcopyright{acmlicensed}\acmConference[KDD '25]{Proceedings of the 31st ACM
SIGKDD Conference on Knowledge Discovery and Data Mining V.2}{August 3--7,
2025}{Toronto, ON, Canada}
\acmBooktitle{Proceedings of the 31st ACM SIGKDD Conference on Knowledge
Discovery and Data Mining V.2 (KDD '25), August 3--7, 2025, Toronto, ON, Canada}
\acmDOI{10.1145/3711896.3737252}
\acmISBN{979-8-4007-1454-2/2025/08}

\begin{document}

\title{NRFormer: Nationwide Nuclear Radiation Forecasting with Spatio-Temporal Transformer}

\author{Tengfei Lyu}
\affiliation{%
  \institution{The Hong Kong University of Science and Technology (Guangzhou)}
  \city{Guangzhou}
  \country{China}}
\email{tlyu077@connect.hkust-gz.edu.cn}

\author{Jindong Han}
\affiliation{%
  \institution{The Hong Kong University of Science and Technology}
  \city{Hong Kong}
  \country{China}}
\email{jhanao@connect.ust.hk}

\author{Hao Liu}
\authornote{Corresponding author.}
\affiliation{%
  \institution{The Hong Kong University of Science and Technology (Guangzhou)}
  \institution{The Hong Kong University of Science and Technology}
  \city{Guangzhou \& Hong Kong}
  \country{China}}
\email{liuh@ust.hk}

\renewcommand{\shortauthors}{Tengfei Lyu, Jindong Han, and Hao Liu.}

\begin{abstract}
Nuclear radiation, which refers to the energy emitted from atomic nuclei during decay, poses significant risks to human health and environmental safety. Recently, advancements in monitoring technology have facilitated the effective recording of nuclear radiation levels and related factors, such as weather conditions. The abundance of monitoring data enables the development of accurate and reliable nuclear radiation forecasting models, which play a crucial role in informing decision-making for individuals and governments. However, this task is challenging due to the imbalanced distribution of monitoring stations over a wide spatial range and the non-stationary radiation variation patterns. In this study, we introduce NRFormer, a novel framework tailored for the nationwide prediction of nuclear radiation variations. By integrating a non-stationary temporal attention module, an imbalance-aware spatial attention module, and a radiation propagation prompting module, NRFormer collectively captures complex spatio-temporal dynamics of nuclear radiation. Extensive experiments on two real-world datasets demonstrate the superiority of our proposed framework against 11 baselines. NRFormer has been deployed online to provide 1–24-day nuclear radiation forecasts, empowering individuals and governments with timely, data-driven decisions for emergency response and public safety. 
Our framework is designed for general applicability and can be readily adapted for deployment in other regions.
The deployed system is available at \url{https://NRFormer.github.io} and the dataset and code of the predictive model are available at \url{https://github.com/usail-hkust/NRFormer}.
\end{abstract}



\maketitle

\section{Introduction}
\label{Introduction}

\begin{figure}[!t]
  \centering
  \includegraphics[width=1\linewidth]{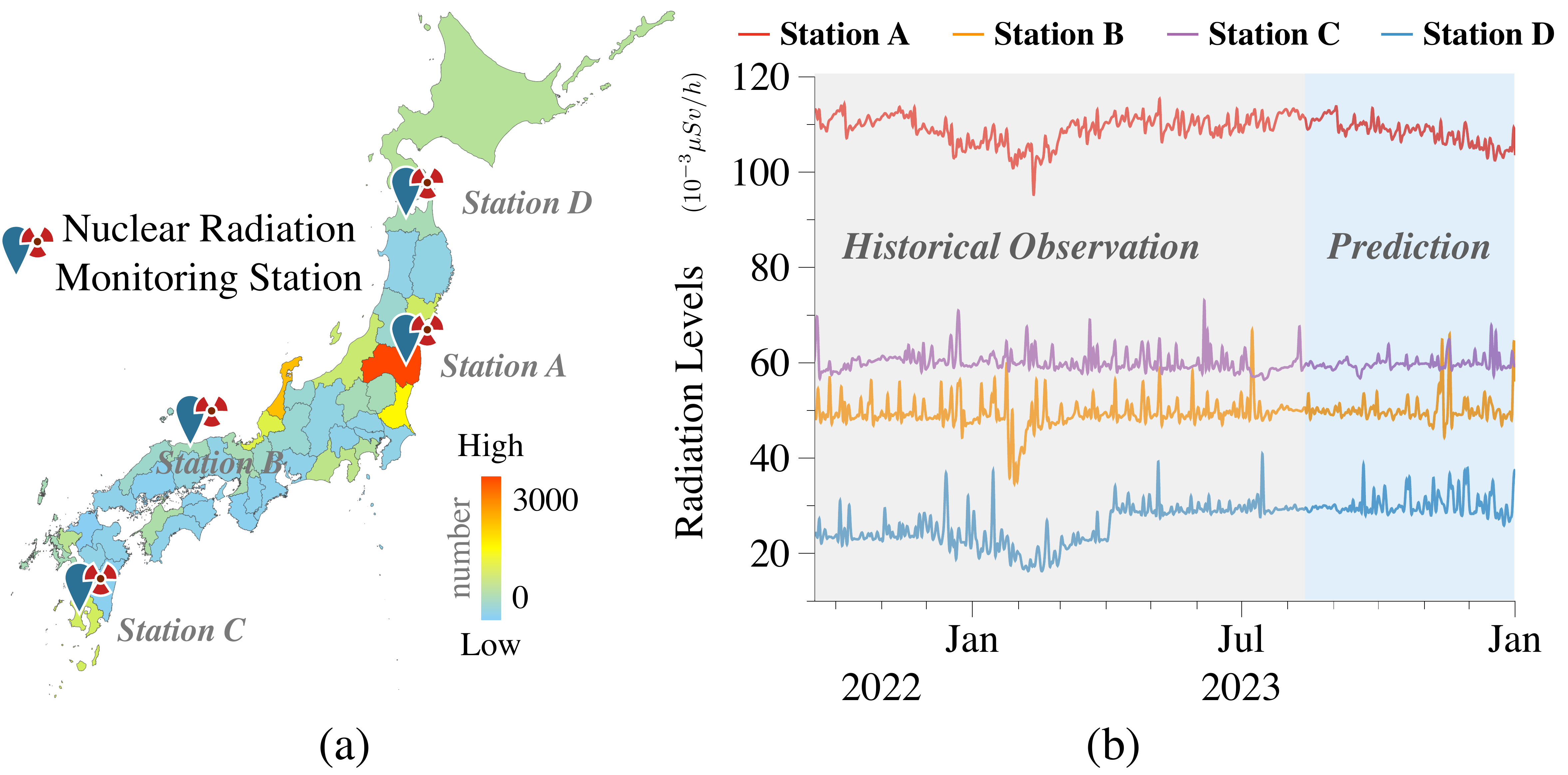}
  \vspace{-0.6cm}
  \caption{Collaborative nuclear radiation forecasting for national-wide stations in Japan. (a) Spatial distribution of monitoring stations in Japan, high-risk areas (\ie with more nuclear power plants) have more stations. (b) Time-varying radiation levels at different monitoring stations.}
  \label{fig:problem-define}
  \Description{}
  \vspace{-0.2cm}
\end{figure}

Nuclear power is recognized as a clean and economical energy source and has been a part of global energy for more than 50 years. However, the potential leakage of nuclear radioactive material poses significant threats to environmental and public safety.  Severe nuclear disasters, such as the Chernobyl accident in the Soviet Union~\cite{ager2019wildfire} and the Fukushima accident in Japan~\cite{steinhauser2014comparison}, have underscored the far-reaching and long-lasting impacts of nuclear radiation across vast geographical areas, affecting diverse ecosystems and public health.
In response, numerous countries have established monitoring stations to track nuclear radiation levels (\eg gamma rays) in the environment. Beyond real-time monitoring, there is an increasing demand for predictive modeling of future radiation level variations. Accurate radiation forecasting can substantially reduce potential socio-economic losses by enabling timely precautionary measures, such as containment of radioactive contamination, evacuation planning, and distribution of protective equipment.

In this paper, we study the national-wide nuclear radiation forecasting problem, which involves collaborative forecasting across distributed monitoring stations throughout the country. 
Figure~\ref{fig:problem-define} shows an illustrative example of radiation forecasting in Japan. 
The dispersion of nuclear radiation is primarily influenced by the global circulation system (\ie atmospheric and ocean circulation)~\cite{qiao2011predicting}, manifesting intricate spatio-temporal patterns. 
Therefore, an effective forecasting model must be capable of comprehending these spatio-temporal correlations from geo-distributed stations. 
Recent advances in Spatio-Temporal Graph Neural Networks (STGNNs)~\cite{jin2023spatio,wu2020connecting,shao2022decoupled,lan2022dstagnn,zhang2022dynamic,shao2022pre}, which combine Graph Neural Networks (GNNs) with sequential models (\eg RNN), have shown superiority in various applications such as traffic forecasting~\cite{wu2019graph,bai2020adaptive,lyu2024autostf,han2024bigst} and air quality prediction~\cite{han2023machine}. Nonetheless, existing STGNNs often fall short in capturing the complex dynamics of nuclear radiation, primarily due to the following three major challenges.

\begin{figure}[t]
  \centering
  \includegraphics[width=1\linewidth]{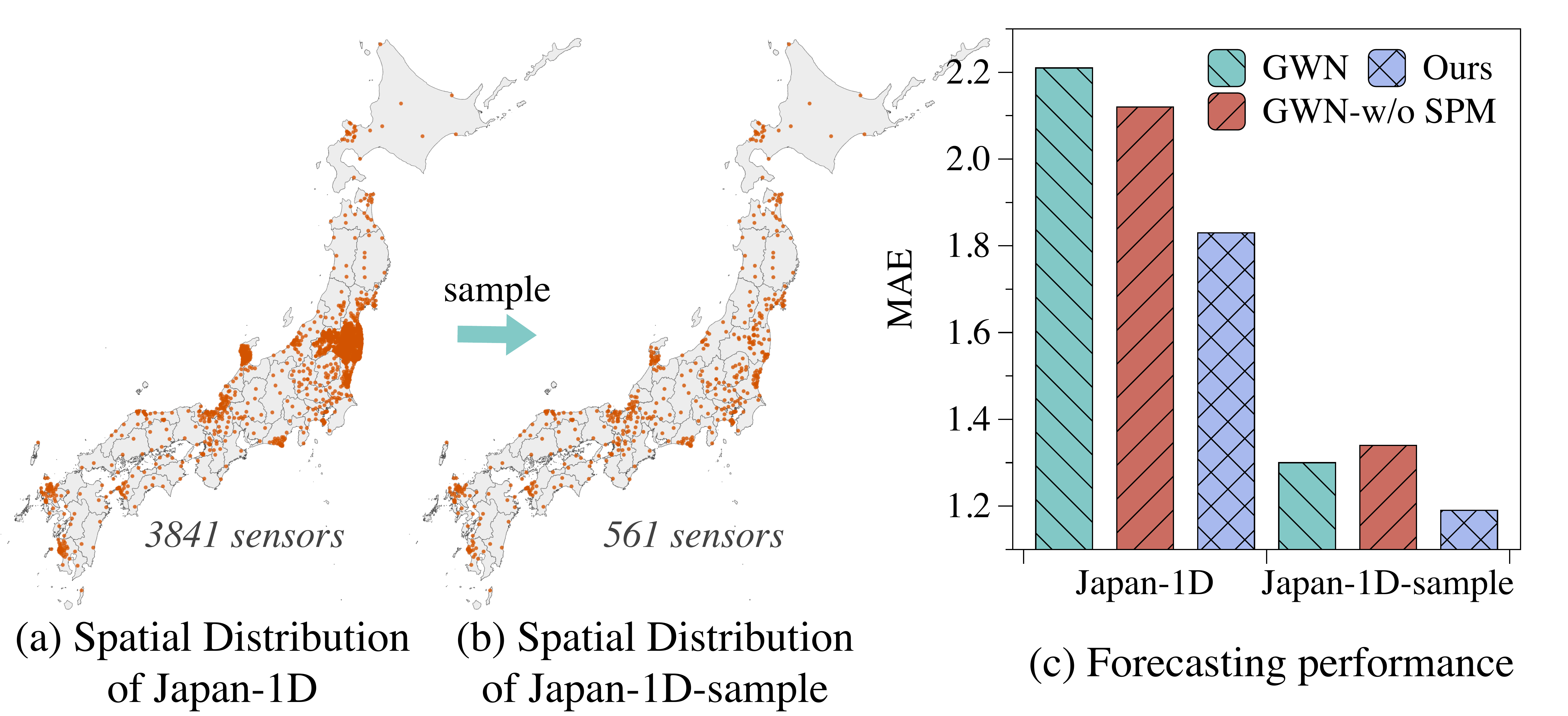}
  \vspace{-0.5cm}
  \caption{
  Empirical study of the imbalanced spatial distribution problem. (1) Full station distribution of our constructed Japan-1D dataset. (2) Evenly sampled station distribution (from Japan-1D dataset). (3) The forecasting performance of GWN on Japan datasets, where GWN-w/o SPM is the model variant that removes the spatial propagation module (SPM).
  }
  \label{fig:introduction}
  \Description{}
  \vspace{-0.5cm}
\end{figure}

\textbf{Challenge 1: Non-stationary temporal patterns.} The nuclear radiation variability is typically driven by unpredictable events like irregular human interventions or nuclear power plant leakages, exhibiting a highly non-stationary nature. This means the statistical properties (\eg mean and variance) and the underlying distribution of radiation series is shifting over time. Traditional STGNNs, which rely on Convolution Neural Networks (CNNs) or Recurrent Neural Networks (RNNs) for temporal pattern preservation~\cite{shao2023exploring}, may struggle to effectively model these erratic, non-stationary properties, thereby affecting predictability. How to mitigate the non-stationary effect towards better predictability is the first challenge.

\textbf{Challenge 2: Imbalanced spatial distribution.} Radiation monitoring stations are often deployed unevenly, with dense clusters near nuclear facilities and sparse distribution in other regions. Typical STGNNs capture spatial correlations by representing these stations as graph nodes and propagating information between adjacent nodes. However, such an approach is less effective in scenarios with spatial imbalances, as nodes in densely populated regions tend to be easily overwhelmed by abundant information, while sparsely connected nodes in other areas fail to perceive sufficient spatial context for prediction. For instance, we illustrate our insights using GWN~\cite{wu2019graph}, a widely adopted spatio-temporal model. As reported in Figure~\ref{fig:introduction}, we observe that removing the spatial propagation module of GWN (\ie GWN-w/o SPM) diminished forecasting effectiveness on a balanced subset, whereas the opposite results are observed when using the full dataset. 
Such results indicate that the utility of the graph learning module is compromised, leading to reduced accuracy in datasets with significant spatial imbalances. Consequently, how to capture diversified dependencies from uneven spatial distribution is also a critical challenge.

\textbf{Challenge 3: Heterogeneous contextual factors.} The variation of radiation at each station is also influenced by a variety of contextual factors, including meteorological conditions, terrain variations, and significant events. These factors may interact with each other, jointly affecting the path and intensity of nuclear radiation. How to effectively incorporate heterogeneous contextual factors is imperative for improving the model's accuracy and reliability.

To address the above challenges, we introduce NRFormer, a spatio-temporal graph Transformer network for national-wide nuclear radiation forecasting.
Our approach is motivated by the recent success of Transformer~\cite{vaswani2017attention} in modeling complex data dependencies through an attention mechanism. 
Specifically, we first devise a non-stationary temporal attention module, which addresses the distribution shift in radiation time series by integrating instance-level normalization into point-wise temporal attention. After that, we propose an imbalance-aware spatial attention module to selectively absorb spatial knowledge for nodes with extremely dense and scarce neighborhoods, thus alleviating the spatial imbalance challenge. 
Moreover, we develop a radiation propagation prompting strategy to guide the predictive modeling process. The prompts not only inject context-specific knowledge into our model for better radiation propagation but also enable robust generalization in unseen scenarios.
Our contributions are summarized as follows:
\begin{itemize}
    \item We investigate the nationwide nuclear radiation forecasting problem. To the best of our knowledge, this is the first work to collectively predict radiation on a national scale.
    \item We propose NRFormer, a predictive model specifically designed to address distinct challenges in nuclear radiation forecasting. By integrating a non-stationary temporal attention module, an imbalance-aware spatial attention module, and a radiation propagation prompting module, NRFormer effectively captures complex spatio-temporal dynamics inherent to nuclear radiation phenomena.
    \item We construct two large-scale nuclear radiation datasets named Japan-4H and Japan-1D, encompassing three years of nuclear radiation data from 3,841 monitoring stations across Japan. We release the dataset \footnote{\url{https://github.com/usail-hkust/NRFormer}} to facilitate future research in this area. Extensive experiments validate the effectiveness of NRFormer against 11 baselines. 
    \item We deploy a web-based, real-time radiation forecasting system \footnote{\url{https://NRFormer.github.io}}, providing 1–24-day predictions and rich radiation data analysis, while offering interactive insights into spatial-temporal radiation patterns.
\end{itemize}

\begin{figure*}[t]
  \centering
  \includegraphics[width=1\linewidth]{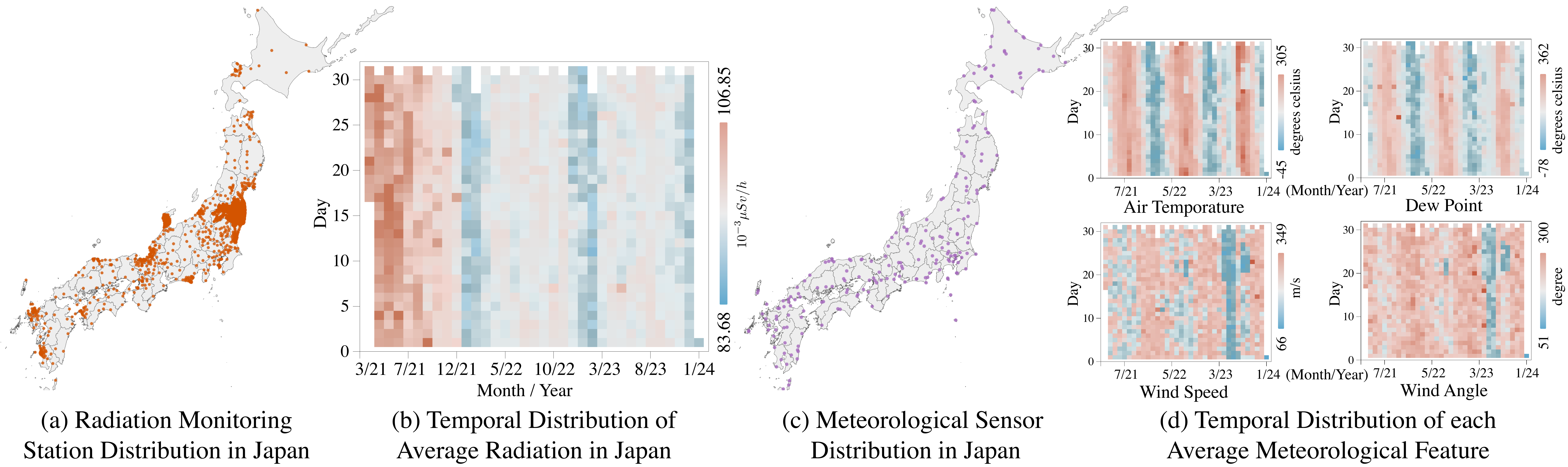}
  \vspace{-0.4cm}
  \caption{Distributions of the radiation and meteorological datasets: (a) spatial distribution of radiation monitoring stations; (b) temporal distribution of daily average radiation in Japan; (c) spatial distribution of meteorological stations; (d) temporal distribution of the daily average value of each meteorological station across Japan.}
  \label{fig:data_analysis}
  \vspace{-0.2cm}
  \Description{}
\end{figure*}

\section{Preliminaries}
\label{sec:Preliminaries}
In this section, we introduce some important definitions and formally define the nuclear radiation forecasting problem.

\begin{myDef}
\textbf{Radiation level}. Radiation level is a number used to quantify the concentrations of radioactive materials in the environment. A higher radiation level indicates that people will experience increasingly detrimental health effects. In practice, the radiation level is computed by using the weighted sum of several radioactive substance measurements, including alpha particles, beta particles, neutron particles, and gamma rays. 
\end{myDef}

\begin{myDef}
\textbf{Radiation monitoring network}. The radiation monitoring network consists of a group of monitoring stations, denoted as $\mathcal{G} = (\mathbf{V}, \mathbf{E})$, where $\mathbf{V}$ is a set of stations and $N=|\mathbf{V}|$ is the number of stations, $\mathbf{E}$ denotes a set of edges representing the relationships among stations. Here we use $\mathbf{A} \in \mathbb{R}^{N \times N}$ to denote the adjacency matrix of the monitoring network.
\end{myDef}

To build $\mathcal{G}$, we compute the pairwise distances between stations and derive the adjacency matrix using a pre-defined distance threshold. Let $\mathbf{X} \in \mathbb{R}^{T \times N}$ be the observed radiation level from all stations, where $T$ is the number of time steps. We use $\mathbf{C} \in \mathbb{R}^{T \times N \times C}$ to denote the contextual features associated with each station, \eg meteorological and location information, where $C$ is the feature dimension. Let $\mathcal{H}^{t} = (\mathcal{G}, \mathbf{X}^{t}, \mathbf{C}^{t})$ indicates all the observed values at time step $t$, we define the problem as follows.

\begin{problem}
\textbf{Nuclear radiation forecasting.}
Given the radiation monitoring network $\mathcal{G}$, historical radiation levels $\mathbf{X}$, contextual features $\mathbf{C}$, the goal is to predict the radiation level for all the monitoring stations over the next $K$ time steps:
\begin{equation}
(\mathbf{\hat{Y}}^{t+1}, \mathbf{\hat{Y}}^{t+2}, \cdots ,\mathbf{\hat{Y}}^{t+K})  \leftarrow \mathcal{F}_{\theta}(\mathcal{H}^{t-P+1}, \mathcal{H}^{t-P+2}, \cdots \mathcal{H}^{t}),
\end{equation}
where $\mathbf{\hat{Y}}^{t+K}$ denotes the predicted value at time step $t+K$, $\mathcal{F}_{\theta}$ is the forecasting model, $P$ and $K$ are the number of historical and future time steps, respectively.
\end{problem}

\section{System Overview}

\subsection{System Architecture}
\begin{figure}[t]
  \centering
  \includegraphics[width=1\linewidth]{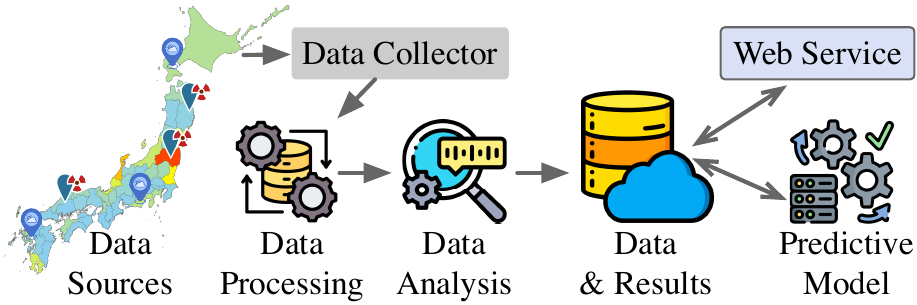}
  \vspace{-0.4cm}
  \caption{Architecture of deployed forecasting system.}
  \label{fig:system}
  \vspace{-0.2cm}
  \Description{}
\end{figure}

Figure \ref{fig:system} illustrates the architecture of our deployed radiation forecasting system. 
The \emph{Data Sources} consist of publicly accessible web services that provide real-time radiation and meteorological data. 
The \emph{Data Collector} systematically acquires time-series data from external sources through web-service APIs or automated web-scraping techniques. 
During the \emph{Data Processing} and \emph{Data Analysis} stages, raw data are cleaned and stored in a cloud-based database. 
The \emph{Predictive Model} leverages this refined dataset to generate forecasts of radiation levels for each radiation monitoring station over the next 1 to 24 days. 
These forecasts are stored in the cloud database and retrieved by the \emph{Web Service}, which provides access to forecasting results and enables interactive visualization and analysis of radiation data through dynamic web interfaces.

\subsection{Data Collecting and Processing}

\textbf{Nuclear radiation data}. The nuclear radiation data are continuously collected from Japan's Nuclear Regulation Authority \footnote{\url{https://www.erms.nsr.go.jp/nra-ramis-webg/}}, consisting of 10-minute interval radiation measurements from over 4,000 monitoring stations.
Stations with more than 30 days of missing data were excluded, resulting in a refined dataset of 3,841 stations.
To analyze radiation forecasting from both long-term and short-term perspectives, the original data were preprocessed to generate two datasets: (1) \textit{Japan-4H}, comprising 4-hour averaged values, and (2) \textit{Japan-1D}, containing daily averaged values.

\textbf{Meteorological data.} Considering the significant influence of meteorological conditions on the spread of radiation, we continuously collect Integrated Surface Dataset (ISD) from the National Oceanic and Atmospheric Administration (NOAA)~\footnote{\url{https://www.ncei.noaa.gov/metadata/geoportal/rest/metadata/item/gov.noaa.ncdc:C00532/html}}. 
The ISD dataset includes hourly surface meteorological observations from over 35,000 weather stations worldwide, covering a variety of parameters like pressure, temperature, dew point, winds, and so on.
By using the geographic coordinates, we identify 235 weather stations in Japan and match them to corresponding radiation monitoring stations. We extract key meteorological variables, including wind speed, wind direction, air temperature, and dew point, as contextual features for radiation forecasting task.

\begin{figure}[t]
  \centering
  \includegraphics[width=1\linewidth]{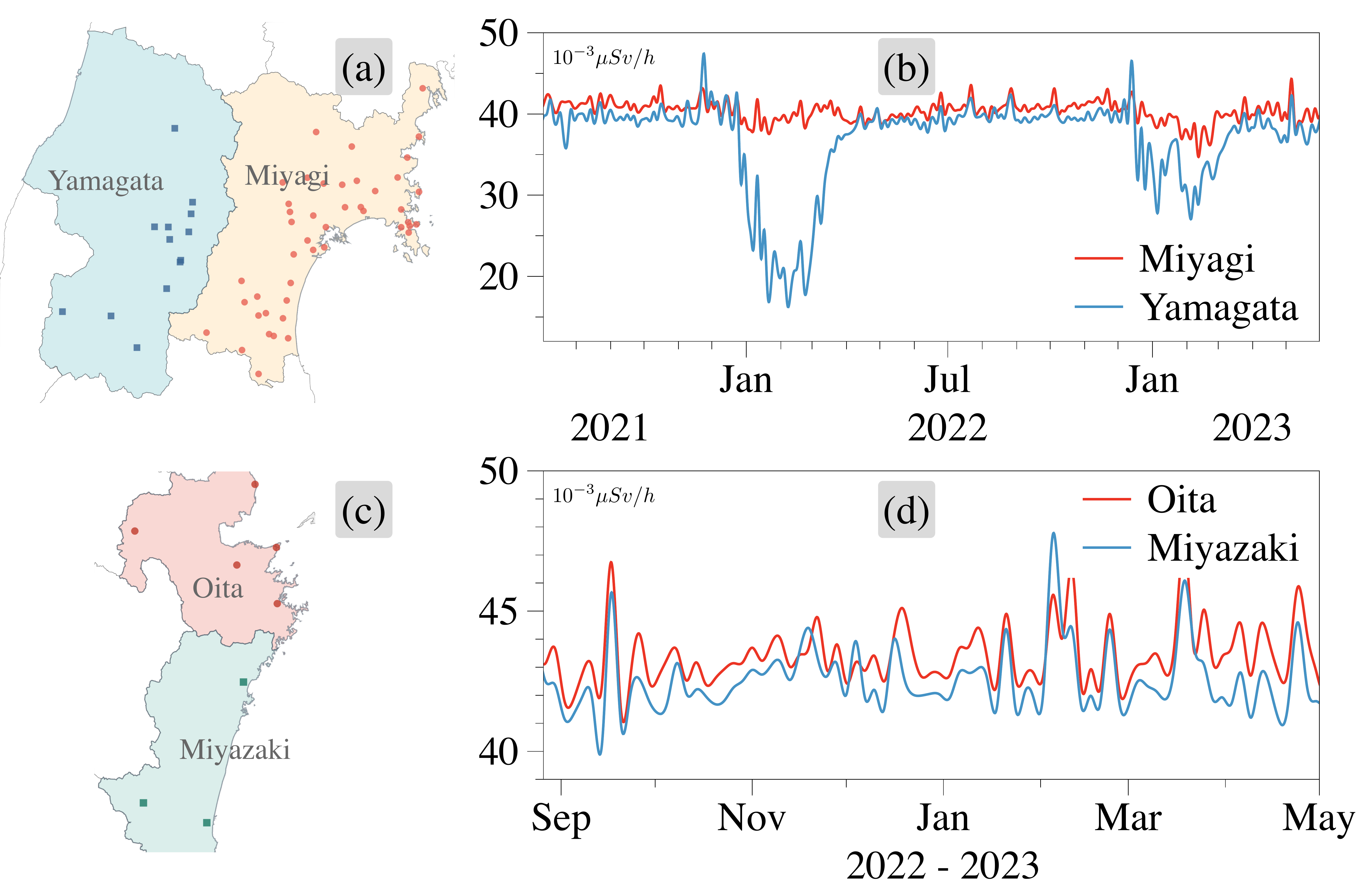}
  \caption{Comparison of radiation patterns in adjacent prefectures, which demonstrates that radiation patterns in neighboring areas can be either very similar or quite dissimilar. (a) and (c) shows the geographical distribution of radiation stations. (b) and (d) present the variation of average radiation levels in corresponding prefectures.}
  \label{fig:data_A2}
  \vspace{-0.4cm}
  \Description{}
\end{figure}

\subsection{Data Analysis}
To investigate radiation distribution patterns, we perform a preliminary analysis of radiation and meteorological data, as illustrated in Figure~\ref{fig:data_analysis}. First, Figures~\ref{fig:data_analysis}(a) and \ref{fig:data_analysis}(c) visualize the spatial distributions of radiation and weather stations, respectively. The analysis reveals a highly skewed spatial distribution of radiation stations, \ie, dense clustering near high-risk nuclear facilities. Figure~\ref{fig:data_analysis}(b) displays the temporal distribution of average radiation levels, showing a pronounced downward trend beginning March 17, 2021, with January exhibiting the lowest annual levels—a pattern potentially linked to seasonal meteorological conditions.

We further analyze spatial and temporal distributions across adjacent prefectures in Figure~\ref{fig:data_A2}, which demonstrates significant non-linear variations in radiation levels across both dimensions. Notably, radiation patterns in geographically proximate regions are not necessarily more similar than those in distant areas, likely due to external complexities such as unpredictable human activities. These dynamics complicate the modeling of spatio-temporal correlations, particularly for densely clustered monitoring stations.

Additionally, we empirically examine relationships between meteorological factors and radiation levels. Figure~\ref{fig:data_analysis}(d) illustrates the temporal distributions of four meteorological variables: wind speed, wind direction, air temperature, and dew point. Temperature and dew point exhibit daily patterns strongly correlated with radiation trends in Figure~\ref{fig:data_analysis}(b). While wind parameters lack explicit correlations, their potential implicit influence on radiation propagation justifies their inclusion. This analysis confirms statistically significant relationships between meteorological conditions and radiation levels, with their integration substantially improving accuracy.

\subsection{User Interfaces}
We developed a web-based real-time forecasting system to provide nationwide radiation level predictions across Japan. The deployed system is presented in Figure~\ref{fig:appendix_forecasting} and~\ref{fig:appendix_data_vis} in the Appendix.

This deployed system enables users to monitor temporal changes in radiation levels. As shown in Figure~\ref{fig:appendix_forecasting}, each monitoring station is represented by a blue marker on the map. Selecting a station triggers a pop-up chart that depicts forecasted radiation levels for subsequent days, offering insights into projected radiation trends. These visualizations assist stakeholders in environmental and public health risk assessment and mitigation planning.
Additionally, Figure~\ref{fig:appendix_forecasting} features a heatmap displaying daily radiation forecasts across Japan. The heatmap highlights geospatial variations in radiation levels, revealing patterns and dispersion dynamics over time. This enables policymakers to identify regional disparities and temporal progression of radiation exposure risks.

As illustrated in Figure~\ref{fig:appendix_data_vis}, the system provides multifaceted radiation analysis through multiple visualization modalities. This system allows both individuals and government agencies to examine radiation data from diverse perspectives, fostering a holistic comprehension of nuclear risks and supporting data-driven decision-making processes.

\begin{figure*}[t]
  \centering
  \includegraphics[width=1\linewidth]{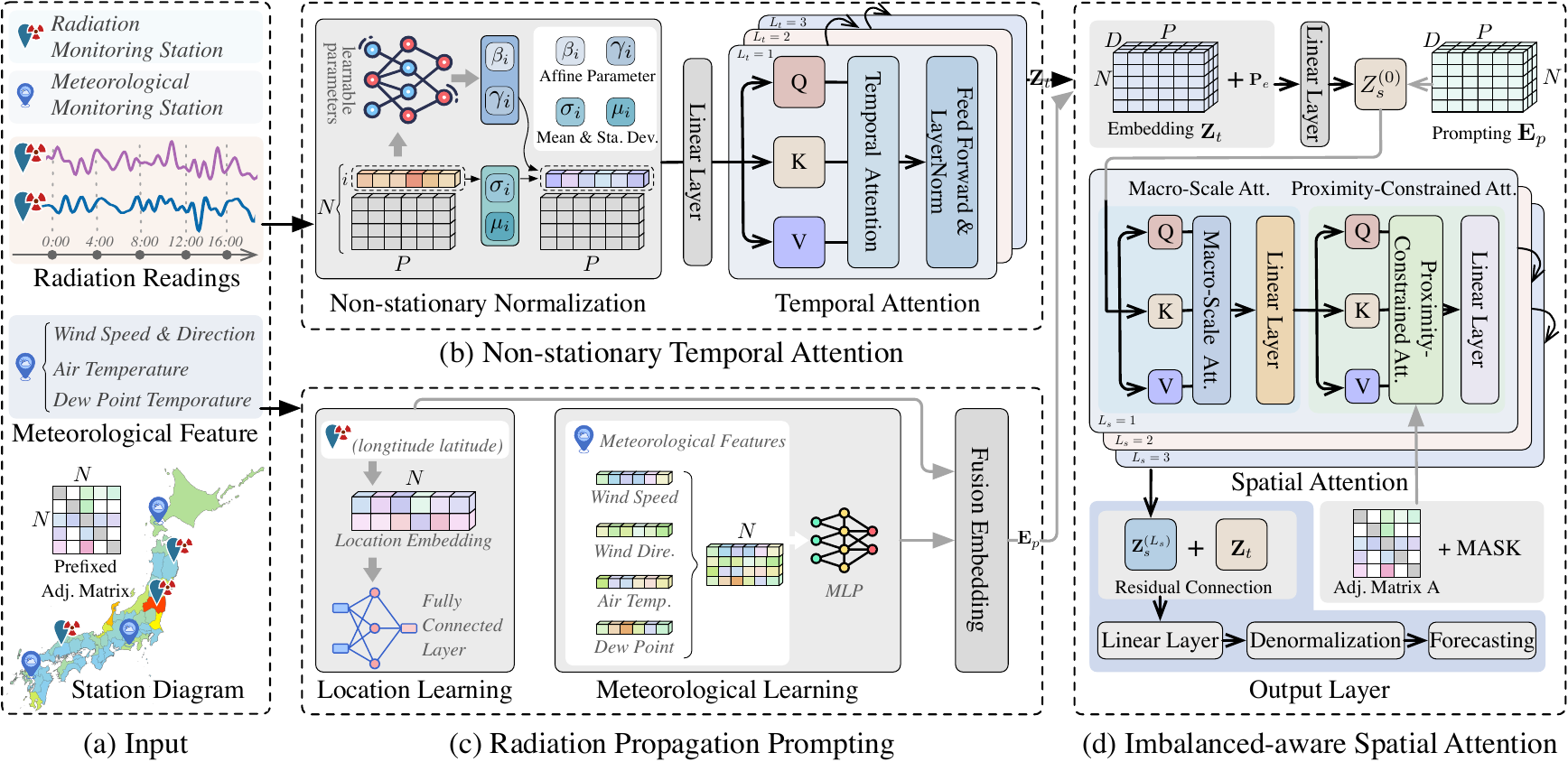}
  \vspace{-0.5cm}
  \caption{The framework overview of NRFormer.}
  \label{fig:framework}
  \Description{}
  \vspace{-0.2cm}
\end{figure*}

\section{The Predictive Model}
\label{sec:The Psroposed Methods}

\textbf{Overview.}
The framework of NRFormer is shown in Figure~\ref{fig:framework}, which is comprised of three modules: 
(1) \textit{The Non-stationary Temporal Attention:} 
it integrates instance-wise normalization into point-wise temporal attention to capture the non-stationary dynamics of radiation. 
(2) \textit{The Imbalance-aware Spatial Attention:}
it models both macro-scale and proximity-constrained spatial dependencies to selectively propagate spatial knowledge for nodes with extremely dense or sparse neighborhoods.
(3) \textit{The Radiation Propagation Prompting:}
it utilizes contextual features to provide a contextual prompt for the imbalanced-aware spatial attention module, enhancing prediction with essential environmental and spatial insights.
Finally, the output layer fuses the embeddings from temporal and spatial attention, and then denormalizes them to produce the final prediction results.
Next, we present each component in detail. 

\subsection{Non-stationary Temporal Attention}
As depicted in Figure 4, the radiation time series often showcase a highly non-stationary nature. To effectively extract stable knowledge from radiation sequence inputs, we first introduce the non-stationary temporal attention, which consists of two parts: non-stationary normalization and point-wise temporal attention.

\subsubsection{Non-stationary normalization}
Due to the non-stationarity of radiation time series, the underlying distributions of different input sequences are diverse, which significantly degrades the forecasting performance. Motivated by recent works~\cite{kim2021reversible,liu2022non}, we aim to eliminate the non-stationary information in each input sequence by using instance-wise normalization.

Formally, given historical observations $\mathbf{x}_{i} \in \mathbb{R}^{P}$ at $i$-th station, where $P$ is the input time window. We first calculate the mean and variance of each input sequence $\mathbf{x}_{i}$ as follows:
\begin{equation}
\begin{aligned}
\mathbb{E}[\mathbf{x}_{i}] = \frac{1}{P}\sum_{j=1}^{P}\mathbf{x}^{j}_{i},\quad Var[\mathbf{x}_{i}] = \frac{1}{P}\sum_{j=1}^{P}(\mathbf{x}^j_{i}-\mathbb{E}[\mathbf{x}_{i}])^{2},
\end{aligned}
\end{equation}
where $\mathbb{E}[\cdot]$ and $Var[\cdot]$ are the mean and variance function, respectively.
After that, we leverage $\mathbb{E}[x_{i}]$ and $Var[x_{i}]$ to normalize the original input sequence $x_{i}$, defined as
\begin{equation}
\begin{aligned}
\hat{\mathbf{x}}_{i} = \gamma_i \frac{\mathbf{x}_{i}-\mathbb{E}[\mathbf{x}_{i}]} {\sqrt{Var[\mathbf{x}_{i}] + \epsilon}} + \beta_i,
\end{aligned}
\end{equation}
where $\gamma_i$ and $\beta_i$ are learnable parameters corresponding to the $i$-th station, and $\epsilon$ is a small value that ensures numerical stability. Intuitively, the normalized sequences possess more stable statistical properties, which largely reduce the difficulty of capturing temporal dynamics within the non-stationary data. 
Subsequently, we process $\hat{\mathbf{x}}_{i}$ through an embedding layer:
$\mathbf{z}_{t}^{i} = Embed(\hat{\mathbf{x}}_{i})$,
where $\mathbf{z}_{t}^{i} \in \mathbb{R}^{P \times D}$ represents the point-wise time series embedding of station $i$ that will be fed into subsequent temporal attention module, $D$ is the hidden dimension of each time step.

\subsubsection{Point-wise temporal attention}
Many STGNNs~\cite{jin2023spatio} rely on CNNs or RNNs to capture the dependencies among different time steps. However, these models may not be able to capture unstable and long-range temporal dependencies, both of which appear in nuclear radiation time series. We address this issue by utilizing a point-wise temporal attention module.

Concretely, we follow the terminology in~\cite{vaswani2017attention} and leverage \emph{query}, \emph{key}, and \emph{value} to indicate the intermediate representation vectors in self-attention mechanism. Given embedding $\mathbf{z}_{t}^{i}$, we derive the \emph{query}, \emph{key}, and \emph{value} matrices by using the following equations:
\begin{equation}
\begin{aligned}
\mathbf{Q}_t = \mathbf{z}_{t}^{i} \mathbf{W}_t^Q,\quad \mathbf{K}_t = \mathbf{z}_{t}^{i} \mathbf{W}_t^K,\quad \mathbf{V}_t = \mathbf{z}_{t}^{i} \mathbf{W}_t^V,
\end{aligned}
\end{equation}
where $\mathbf{W}_t^Q, \mathbf{W}_t^K, \mathbf{W}_t^V \in \mathbb{R}^{D\times D}$ are learnable parameters shared across different stations and time steps.
Afterward, we compute attention score $A_t$ to measure relationships among any two time steps, 
$\mathbf{A}_t = \frac{\mathbf{Q}_t {\mathbf{K}_t}^{\top}}{\sqrt{D}}$,
where $A_t \in \mathbb{R}^{P\times P}$ and $\sqrt{D}$ is the scaling factor.

\eat{where $A_t \in \mathbb{R}^{N\times P\times P}$ captures the temporal relations across various spatial nodes, effectively mapping the interconnectedness of data points over time.}

\eat{The output of the temporal attention denoted as $Z^o_t \in \mathbb{R}^{N\times P\times D_h}$, is then derived by combining the self-attention scores with the value matrix:}
We normalize each row of $\mathbf{A}_t$ with a Softmax function, and obtain the updated representation vector as follows
$\mathbf{z}_{t}^{i,(l)} = Softmax(\mathbf{A}_t) \mathbf{V}_t$.
To enhance the effectiveness of temporal dependency modeling, we further equip multi-head attention and residual connection technique for each layer, followed by layer normalization~\cite{vaswani2017attention}. We build the temporal attention module by stacking $L_t$ such layers, and denote the output of the $L_t$-th layer as $\mathbf{z}_{t}^{i,(L_t)}$. 

Finally, we flatten $\mathbf{z}_{t}^{i,(L_t)}$ into a 1D vector and map the flattened version into a hidden representation,
$\mathbf{z}_{t}^{i} = Flatten(\mathbf{z}_{t}^{i,(L_t)}) \mathbf{W}^{r}$,
where $\mathbf{W}^{r}$ is learnable parameters. Let $\mathbf{Z}_t \in \mathbb{R}^{N\times D}$ denotes the hidden representations of all the nodes (\ie stations), where the $i$-th row vector of $\mathbf{Z}_t$ is $\mathbf{z}_{t}^{i}$.

\subsection{Imbalance-aware Spatial Attention}
As previously mentioned, the nodes in dense clusters have the risk of information overload (\ie over-smoothing) and sparsely connected nodes have the risk of information insufficient (\ie under-smoothing) in spatial dependency modeling, leading to sub-optimal performance. To alleviate this problem, we develop the imbalance-aware spatial attention, which includes two parts: (1) a macro-scale spatial correlation modeling module to enrich the spatial neighborhood of orphan nodes, and (2) a proximity-constrained spatial modeling module to reduce the influence of noisy neighborhood for densely connected nodes.

\eat{Given $Z^{(K)}_t \in \mathbb{R}^{N\times P\times D}$ generated by the non-stationary temporal attention, we first introduce a positional encoding mechanism~\cite{vaswani2017attention,lai2023lightcts} to the embedding $Z_t^o$. 
The reason is that by default, Transformers’ self-attention operations are unable to interpret the permutations of the input nodes.
Specifically, the encoding layer converts $Z_t^o$ to a learned positional encoding $W^p$ by incorporating the identity information of radiation stations:}

Following previous works~\cite{vaswani2017attention,lai2023lightcts}, we add positional encoding into $\mathbf{Z}_s$ to preserve critical spatial information of each node:
$\mathbf{Z}_s^{(0)} = \mathbf{Z}_t + \mathbf{P}_e$,
where $\mathbf{P}_e \in \mathbb{R}^{N\times D}$ is a learnable positional embedding matrix. The resulting matrix $\mathbf{Z}_s^{(0)}$ will be fed into the following spatial attention block for spatial dependency modeling.

\eat{where $W^{p} \in \mathbb{R}^{N\times D}$ is a learnable matrix capturing the identity information. 
The resulting encoding $Z_s^{(0)}$ is fed to sequential spatial attenuation blocks, each of which consists of a multi-head attention layer and a feed-forward network layer.}

\subsubsection{Macro-Scale Spatial Correlation Modeling}
\eat{We first introduce the multi-attention mechanism, which is a concatenation of $t$ self-attention heads to capture macro-scale spatial correlations among all radiation stations. 
Specifically, given attention head $h$, the learnable matrices~$W_s^Q, W_s^K, W_s^V \in \mathbb{R}^{D\times D}$, the attention function can be written as:}

In this part, we employ the macro-scale spatial attention mechanism to identify latent neighborhoods for sparsely-connected nodes. Specifically, we first compute the attention score between arbitrary nodes:
\begin{equation}
\begin{aligned}
\label{equ:global_QKV}
\mathbf{Q}_{g} = \mathbf{Z}_s^{(0)} \mathbf{W}_{g}^Q,\quad \mathbf{K}_{g} = \mathbf{Z}_s^{(0)} \mathbf{W}_{g}^K,\quad \mathbf{V}_{g} = \mathbf{Z}_s^{(0)} \mathbf{W}_{g}^V,
\end{aligned}
\end{equation}
\begin{equation}
\begin{aligned}
\label{equ:global_A}
\mathbf{A}_{g} = \frac{\mathbf{Q}_{g} \cdot \mathbf{K}_{g}^{\top}}{\sqrt{D}},
\end{aligned}
\end{equation}
where $\mathbf{A}_{g} \in \mathbb{R}^{N\times N}$ is the attention score matrix that encodes the all-pair node relationships.
Then, we can propagate node signals based on the normalized $\mathbf{A}_{g}$, defined as
$\mathbf{H}_{g} = Softmax(\mathbf{A}_{g}) \cdot \mathbf{V}_{g}$.
\eat{where $h$ denotes the number of heads. 
After this, the embedding $h_{g}$ is fed into linear layers, which outputs the final embedding calculated by this macro-scale spatial attention at this layer. 
The formal formulation is as follows:}
Next, we further adopt a feed-forward layer to enhance the model expressive capabilities, 
$\mathbf{H}^{\prime}_{g} = (ReLU(\mathbf{H}_{g}\cdot \mathbf{W}_1 + \mathbf{b}_1)) \cdot \mathbf{W}_2 + \mathbf{b}_2$,
where $\mathbf{W}_1, \mathbf{W}_2$ are the learnable matrices and $\mathbf{b}_1, \mathbf{b}_2$ are the bias parameters. In summary, macro-scale spatial attention can provide additional spatial information for nodes with scarce neighborhoods, which reduces the risk of under-smoothing phenomena.

\subsubsection{Proximity-Constrained Spatial Correlation Modeling}

In practice, the macro-scale spatial attention may introduce massive noises for densely-connected nodes via all-pair message passing, which exacerbates the over-smoothing issue. Therefore, we devise proximity spatial attention by adding a graph constraint to filter those neighbors that are far away from the target nodes. Likewise, we first compute the attention score matrix $\mathbf{A}_{l}$ and the \emph{key} matrix $\mathbf{V}_{l}$ via Equations \ref{equ:global_QKV} and \ref{equ:global_A}. Suppose $\mathbf{A}$ denotes the truncated adjacency matrix built from geographical distance, we define a masked matrix $\mathbf{A}_{mask}$ based on $\mathbf{A}$ and $\mathbf{A}_{l}$:
\begin{equation}
\begin{aligned}
\mathbf{A}_{mask}[i,j] = 
\begin{cases} 
  \mathbf{A}_{l}[i,j], & \text{if}\ \mathbf{A}[i,j] > 0 \\
  -\infty, & \text{otherwise} 
\end{cases}
\end{aligned}
\end{equation}
where $\mathbf{A}_{mask}[i,j]$ is set to $-\infty$ if the distance between node $i$ and $i$ exceeds a threshold.
Afterwards, we perform spatial message passing under proximity constraint:
$\mathbf{H}_{l} = Softmax(\mathbf{A}_{mask}) \cdot \mathbf{V}_{l}$.
The output $\mathbf{H}_{l}$ is then fed into a feed-forward layer,
\begin{equation}
\begin{aligned}
\mathbf{H}^{\prime}_{l} = (ReLU(\mathbf{H}_{l}\cdot \mathbf{W}_3 + \mathbf{b}_3)) \cdot \mathbf{W}_4 + \mathbf{b}_4,
\end{aligned}
\end{equation}
where $\mathbf{W}_3, \mathbf{W}_4$ are the learnable matrices and $\mathbf{b}_3, \mathbf{b}_4$ are the bias parameters. The proximity spatial message passing operator can restrict the spatial receptive field of densely connected nodes and enable the adaptive selection of useful neighboring nodes, which can avoid potential over-smoothing risk.
We alternately stack $L_s$ macro-scale and proximity spatial attention layers for spatial dependency modeling. The final output is denoted as $\mathbf{Z}_s^{(L_s)}$.

\subsection{Radiation Propagation Prompting}
\eat{In this section, we aim to enhance the learning of radiation propagation patterns by incorporating contextual features such as geographical features and meteorological features. 
The integrated contextual features serve as a contextual prompt for the imbalanced-aware spatial attention module, enriching it with critical environmental and spatial insights. 
This enhancement allows the model to more effectively account for the complex interplay between geographical location and meteorological conditions in predicting radiation dispersion.}

Considering the significant impact of context factors on radiation propagation, we encode context-specific knowledge as prompts to guide model forecasting. In particular, we incorporate two types of prompts: the location prompt and the meteorological prompt.

\textbf{Location prompt}. Different locations in geographical space often exhibit distinct spatial characteristics, which could potentially affect the spread of radioactive materials. To implicitly capture this critical feature, we introduce the geographical coordinates of each radiation monitoring station, denoted as $\mathbf{L} \in \mathbb{R}^{N\times C_g}$, where $N$ represents the number of stations and $C_g=2$ is feature dimension, \ie longitude and latitude. We project $\mathbf{L}$ into a latent embedding space via an embedding layer, and apply Multi-Layer Perceptron (MLP) to fuse the embeddings, defined as $\mathbf{E}_{l} = MLP_l(Embed(\mathbf{L}))$.

\textbf{Meteorological prompt}. Meteorological conditions significantly influence the spatial propagation of nuclear radiation. For example, strong wind can carry radioactive material over long distances, affecting areas far from the contamination source. Therefore, for each radiation monitoring station, we retrieve the observations of its nearest meteorological station as prompt features, denoted as $\mathbf{M} \in \mathbb{R}^{N\times T\times C_m}$, where $C_m$ is the number of meteorological features. We flatten the last two dimensions of $\mathbf{M}$ into a 2D matrix $\mathbf{M}^{\prime} \in \mathbb{R}^{N\times C_o}$, where $C_o=T\times C_m$. Similarly, we transform the raw features into a latent feature representation through an MLP: $\mathbf{E}_{m} = MLP_m(\mathbf{M}^{\prime})$.
After that, we fuse the above two prompt feature vectors $\mathbf{E}_p = \mathbf{E}_{l} || \mathbf{E}_{m}$, where $\mathbf{E}_p$ is the contextual prompt embedding that will be injected into the imbalanced-aware spatial attention as addition clues. Concretely, we can rewrite Equation \ref{equ:global_QKV} as 
\begin{equation}
\begin{aligned}
{\mathbf{Q}}_{g} = (\mathbf{Z}_s^{(0)} ||\mathbf{E}_p)\mathbf{W}_g^Q, \mathbf{K}_{g} = (\mathbf{Z}_s^{(0)}||\mathbf{E}_p)\mathbf{W}_g^K,  \mathbf{V}_{g} = \mathbf{Z}_s^{(0)}\mathbf{W}_g^V.
\end{aligned}
\end{equation}
Overall, the prompting strategy enables our model to take the complex interplay between geographical location and meteorology into account in radiation propagation modeling.

\eat{Specifically, we collected the geographical coordinates of each radiation monitoring station, denoted as $\mathbf{L} \in \mathbb{R}^{N\times C_g}$, where $C_g=2$ denotes the longitude and latitude. Additionally, for each radiation monitoring station, we selected the nearest one from 235 meteorological monitoring stations to represent the meteorological characteristics of that radiation station, denoted as $\mathbf{M} \in \mathbb{R}^{T\times N\times C_m}$, where $C_m$ is the number of meteorological features.

We use the fully connected layers to extract the features of $\mathbf{L}$ and $\mathbf{L}$, respectively:
\begin{equation}
\begin{aligned}
E_{l} = MLP_l(\mathbf{L}),\quad E_{m} = MLP_m(\mathbf{M}),\quad Z_p = concat[E_{l}, E_{m}],
\end{aligned}
\end{equation}
where $E_{l}\in \mathbb{R}^{N\times D}$ and $E_{l}\in \mathbb{R}^{N\times D}$ are hidden embedding. $Z_p$ is the contextual feature prompting embedding and will feed into the imbalanced-aware spatial attention module to enhance the radiation forecasting. }

\begin{table*}[ht]
\small
\renewcommand{\arraystretch}{0.85} 
\setlength{\tabcolsep}{2.5pt} 
  \caption{Evaluations of NRFormer and baselines on two real-world datasets. 
  }
  \label{tab:overall results}
  \vspace{-0.3cm}
  \begin{tabular}{c|c|ccc|ccc|ccc|ccc|ccc}
    \toprule

    \multirow{2}{*}{\text { \textbf{Data} }} & \multirow{2}{*}{\text { \textbf{Models} }} & \multicolumn{3}{c|}{ \textbf{6th} (24 hours / 6 days)} & \multicolumn{3}{c|}{ \textbf{9th} (36 hours / 9 days)} & \multicolumn{3}{c|}{ \textbf{12th} (48 hours / 12 days)} & \multicolumn{3}{c|}{ \textbf{24th} (96 hours / 24 days)} & \multicolumn{3}{c}{ \textbf{sudden change}}\\

    & & \text { MAE } & \text { RMSE } & \text { MAPE } & \text { MAE } & \text { RMSE } & \text { MAPE } & \text { MAE } & \text { RMSE } & \text { MAPE } & \text { MAE } & \text { RMSE } & \text { MAPE } & \text { MAE } & \text { RMSE } & \text { MAPE } \\
    
    \midrule

    \multirow{12}{*}{\rotatebox{90}{ Japan-4H }}
    & \text{ HA }            & 3.20 & 19.73 & 3.18\% & 3.57 & 20.16 & 3.58\% & 3.51 & 20.20 & 3.48\%  & 3.74 & 20.78 & 3.64\% & 3.83 & 21.91 & 3.67\% \\
    & \text{ LR }            & 2.80 & 11.17 & 2.58\% & 2.96 & 11.50 & 2.75\% & 2.94 & 11.75 & 2.64\%  & 3.04 & 12.43 & 2.65\% & 3.44 & 14.19 & 2.88\% \\
    & \text{ XGBoost }            & 2.38 & 12.47 & 2.53\% & 2.49 & 12.92 & 2.61\% & 2.58 & 13.10 & 2.67\%  & 2.80 & 14.38 & 2.83\% & 3.64 & 16.46 & 2.97\% \\
    
    & \text{ DCRNN }         & 2.15 & 6.32 & 2.52\% & 2.30 & 7.22 & 2.64\% & 2.42 & 8.01 & 2.73\%  & 2.79 & 10.49 & 2.97\% & 3.09 & 12.12 & 2.81\% \\
    & \text{ STID }          & 2.04 & 5.72 & 2.47\% & 2.15 & 6.18 & 2.56\% & 2.19 & 6.50 & 2.59\%  & 2.30 & 7.51 & 2.63\%  & 2.56 & 9.63 & 2.79\% \\
    & \text{ DLinear }    & 2.12 & 6.16 & 2.57\% & 2.20 & 6.15 & 2.66\% & 2.22 & 6.51 & 2.60\%   & 2.31 & \underline{7.47} & 2.66\% & 2.35 & 8.99 & 2.62\% \\
    & \text{ PatchTST }    & 1.89 & 5.61 & 2.24\% & 2.01 & 6.06 & 2.35\% & 2.05 & 6.36 & 2.36\%   & \underline{2.23} & 7.57 & \underline{2.48\%} & 2.32 & \underline{7.84} & \underline{2.47\%} \\
    & \text{ Koopa }      & 1.92 & \underline{5.69} & \underline{2.28\%} & \underline{2.01} & \underline{6.08} & \underline{2.35\%} & \underline{2.06} & \underline{6.42} & \underline{2.38\%}   & \underline{2.23} & 7.64 & 2.55\% & \underline{2.31} & 7.89 & 2.49\% \\

    & \text{ StemGNN }    & 2.01 & 6.11 & 2.72\% & 2.22 & 6.19 & 2.52\% & 2.29 & 6.47 & 2.55\%  & 2.38 & 7.53 & 2.59\% & 2.41 & 8.71 & 2.71\% \\
    & \text{ GWN } & 2.25 & 5.99 & 2.78\% & 2.26 & 6.30 & 2.72\% & 2.27 & 6.48 & 2.70\%  & 2.40 & 7.33 & 2.80\%  & 2.69 & 9.86 & 2.72\% \\
    & \text{ LightCTS }      & 1.87 & 5.73 & 2.34\% & \underline{2.01} & 6.35 & 2.44\% & 2.15 & 6.54 & 2.49\%   & 2.29 & 7.51 & 2.50\% & 2.40 & 8.69 & 2.68\% \\

    & \textbf{ NRFormer }       & \textbf{1.72} & \textbf{4.82} & \textbf{2.07\%} & \textbf{1.79} & \textbf{5.15} & \textbf{2.14\%} & \textbf{1.85} & \textbf{5.39} & \textbf{2.18\%} &  \textbf{1.99} & \textbf{6.24} & \textbf{2.31\%} &  \textbf{2.05} & \textbf{6.76} & \textbf{2.31\%}\\
    
    \midrule
    \multirow{12}{*}{\rotatebox{90}{Japan-1D }}
    & \text{ HA }            & 2.86 & 17.83 & 2.73\% & 2.97 & 18.22 & 2.86\% & 3.08 & 18.58 & 2.98\%  & 3.46 & 20.01 & 3.45\% & 4.11 & 22.41 & 3.49\% \\
    & \text{ LR }            & 2.46 & 11.08 & 2.07\% & 2.55 & 11.76 & 2.14\% & 2.67 & 12.27 & 2.25\%  & 2.95 & 14.21 & 2.50\% & 3.79 & 13.92 & 3.01\% \\
    & \text{ XGBoost }            & 2.51 & 13.71 & 2.66\% & 2.64 & 14.97 & 2.76\% & 2.73 & 15.36 & 2.85\%  & 2.95 & 16.10 & 3.07\% & 3.82 & 15.14 & 3.15\%\\
    
    & \text{ DCRNN }         & 1.97 & 10.51 & 2.00\% & 2.14 & 11.44 & 2.09\% & 2.78 & 12.43 & 2.17\%  & 2.64 & 14.53 & 2.42\% & 3.46 & 13.63 & 3.02\% \\
    & \text{ STID }          & 2.44 & 7.42 & 3.41\% & 2.57 & 7.93 & 3.58\% & 2.68 & 8.38 & 3.72\%  & 3.06 & 9.99 & 4.23\%  & 3.26 & 12.62 & 3.12\% \\
    & \text{ DLinear }    & 1.89 & 7.85 & 2.03\% & 2.11 & 8.75 & 2.25\% & 2.19 & 8.42 & 2.42\%   & 2.33 & 9.19 & 2.46\%  & 2.48 & 10.54 & 2.60\% \\
    & \text{ PatchTST }      & 1.84 & \underline{6.37} & \underline{1.95\%} & 1.97 & 6.98 & 2.05\% & 2.03 & 7.90 & \underline{2.10\%}  & 2.17 & \underline{9.02} & 2.44\% & 2.44 & \underline{9.92} & 2.53\% \\
    & \text{ Koopa }      & \underline{1.74} & 7.03 & 2.00\% & \underline{1.83} & 7.68 & \underline{2.10\%} & 1.91 & 8.17 & 2.15\%  & \underline{2.11} & 9.77 & 2.49\% & 2.46 & 9.95 & \underline{2.50\%} \\
    
    & \text{ StemGNN }    & 1.83 & 7.79 & 2.11\% & 2.14 & 8.63 & 2.23\% & 2.16 & 8.39 & 2.42\%   & 2.32 & 9.13 & 2.43\%  & 2.42 & 10.25 & 2.56\% \\
    & \text{ GWN } & 1.90 & 7.31 & 2.16\% & 1.97 & 7.77 & 2.19\% & 2.04 & 8.15 & 2.26\%  & 2.20 & 9.19 & 2.43\% & 2.94 & 11.35 & 2.86\% \\
    & \text{ LightCTS }      & 1.85 & 8.02 & 2.13\% & 1.93 & \underline{7.17} & 2.17\% & 2.00 & \underline{7.74} & 2.15\%  & 2.15 & 9.19 & \underline{2.36\%} & \underline{2.39} & 9.97 & 2.51\%\\
    
    & \textbf{ NRFormer }       & \textbf{1.71} & \textbf{6.25} & \textbf{1.90\%} & \textbf{1.78} & \textbf{6.82} & \textbf{1.94\%} & \textbf{1.83} & \textbf{7.28} & \textbf{1.98\%}   & \textbf{2.01} & \textbf{8.76} & \textbf{2.13\%} & \textbf{2.06} & \textbf{8.97} & \textbf{2.21\%} \\

    \bottomrule
  \end{tabular}
  \vspace{-0.2cm}
\end{table*}

\begin{table}[t] \small
  \renewcommand{\arraystretch}{0.75} 
  \caption{Statistic of radiation and meteorological datasets.}
  \label{tab:dataset}
  \vspace{-0.3cm}
  \begin{tabular}{cc|cc}
    \toprule
    \multicolumn{2}{c|}{Data Description} & Japan-4H & Japan-1D \\
    \midrule
    \multirow{4}{*}{\text { Nuclear radiation data }} 
    & \# of stations & 3,841 & 3,841 \\
    & \# of timesteps & 6,121 & 1,021 \\
    & Interval & 4 hour & 1 day \\
    & Time span & \multicolumn{2}{c}{3/17/2021 - 1/1/2024} \\
    
    \midrule
    \multirow{4}{*}{\text { Meteorological data }} 
    & \# of stations & \multicolumn{2}{c}{235} \\
    & \# of timesteps &  \multicolumn{2}{c}{24,483} \\
    & Interval &  \multicolumn{2}{c}{1 hour} \\\
    & Time span &  \multicolumn{2}{c}{3/17/2021 - 1/1/2024} \\
    \bottomrule
  \end{tabular}
  \vspace{-0.2cm}
\end{table}

\subsection{Output Layer}
To make predictions, we concatenate the embedding $\mathbf{Z}_{t}$ and $\mathbf{Z}_{s}^{(L_s)}$, and then feed it into fully connected layers. The formulation is shown as follows:
\begin{equation}
\begin{aligned}
\tilde{\mathbf{Y}}^{t+1:t+K}= ((\mathbf{Z}_{t}||\mathbf{Z}_{s}^{(L_s)}) \cdot W_i+b_i)\cdot W_o + b_o,
\end{aligned}
\end{equation}
where $W_i$ and $W_o$ are the learnable matrices, and $b_i$ and $b_o$ are the bias parameters. 
Suppose $\tilde{\mathbf{y}}^{i} \in \mathbb{R}^{K}$ represents the $i$-th row vector of $\tilde{\mathbf{Y}}^{t+1:t+K}$, we denormalize the model output $\tilde{\mathbf{y}}_{i}$ using the reciprocal of the instance normalization, denoted as
\begin{equation}
\begin{aligned}
\hat{\mathbf{y}}_{i}= \sqrt{Var[\mathbf{x}^{(i)}] + \epsilon}\cdot \frac{\tilde{\mathbf{y}}_{i}-\beta_i}{\gamma_k} + \mathbb{E}[\mathbf{x}^{(i)}],
\end{aligned}
\end{equation}
where $\hat{\mathbf{y}}_{i}$ denotes the future prediction of station $i$.
Finally, we aim to optimize the following objective:
$\mathcal{L}=\frac{1}{N}\sum_{i=1}^{N} |\hat{\mathbf{y}}_{i}-\mathbf{y}_{i}|$,
where $\mathbf{y}_{i}$ is the ground truth radiation level of station $i$.

\section{Experiments}
\label{sec:Experiments}
In this section, we introduce the experiment setup, overall performance, ablation study, parameter sensitivity analysis, and deployed system evaluation.

\subsection{Experiments Setup}

\subsubsection{Datasets}
The experimental datasets for this predictive model include nuclear radiation data and meteorological data collected from the entire country of Japan. 
Table~\ref{tab:dataset} summarizes the statistics of the radiation and meteorological datasets.

\subsubsection{Evaluation Metrics}
We employed three widely used metrics, including Mean Absolute Error (MAE), Root Mean Squared Error (RMSE), and Mean Absolute Percentage Error (MAPE), for model evaluation. Lower values in MAE, RMSE, and MAPE indicate higher forecasting accuracy. 
Moreover, we follow \cite{zheng2015forecasting,yi2018deep,liang2023airformer} to discuss the errors on predicting sudden changes. We identified the top 10\% of samples in the test set with the greatest variation between observed and predicted values as sudden change samples for prediction.

\subsubsection{Baseline}
We compare the proposed NRFormer with 11 baseline methods, including:

\begin{itemize}
\item HA \cite{zhang2017deep}: 
A method predicts nuclear radiation levels by calculating the average value of historical readings for corresponding periods.

\item XGBoost \cite{chen2016xgboost}: 
A method predicts nuclear radiation levels by leveraging an ensemble of decision trees, optimizing for both efficiency and performance. 

\item LR: 
A linear regression model that leverages the linear relationship between input variables and the output.

\item DCRNN \cite{li2018diffusion}: 
a spatio-temporal forecasting model that captures spatial dependencies through bidirectional random walks, while addressing temporal dependencies using encoder-decoder architecture with scheduled sampling.

\item GWN \cite{wu2019graph}: 
A spatial-temporal graph modeling framework that generates an adaptive graph and integrates diffusion graph convolution with dilated causal convolution.

\item STID \cite{shao2022spatial}: 
A simple yet effective model for time series forecasting that only considers the historical time series with spatial-temporal identity embeddings.

\item DLinear \cite{zeng2023transformers}: 
This study introduces a surprisingly simple yet effective one-layer linear model, LTSF-Linear, which outperforms sophisticated Transformer-based long-term time series forecasting models across multiple datasets, challenging the prevailing reliance on complex Transformer architectures for temporal data analysis.

\item StemGNN \cite{cao2020spectral}: 
StemGNN is a novel framework that captures both inter-series correlations and temporal dependencies within multivariate time-series data in the spectral domain.

\item LightCTS \cite{lai2023lightcts}: 
LightCTS is a framework aimed at efficient, lightweight forecasting for correlated time series, balancing accuracy with reduced computational needs. 

\item PatchTST \cite{nietime}: 
PatchTST is a Transformer-based model for multivariate time series forecasting, leveraging subseries patching and channel independence to reduce complexity and improve long-term accuracy.

\item Koopa \cite{liu2024koopa}: 
Koopa introduces a novel approach for forecasting non-stationary time series by leveraging modern Koopman theory to disentangle and advance time-variant and time-invariant dynamics.

\end{itemize}

\begin{table}[t] \small
  \renewcommand{\arraystretch}{0.8} 
  \caption{Ablation studies of NRFormer on two datasets.}
  \vspace{-0.3cm}
  \label{tab:ablation_studies}
  \begin{tabular}{c|c|ccc}
    \toprule

    \text { Dataset } & \text { Models } & \text { MAE } & \text { RMSE } & \text { MAPE }\\

    \midrule
    \multirow{5}{*}{Japan-4H}
    & \text{ w/o Normalization }         & 2.16 & 6.68 & 2.53\% \\
    & \text{ w/o Temporal Attention }    & 2.21 & 7.43 & 2.48\% \\
    & \text{ w/o Spatial Attention }     & 2.24 & 7.36 & 2.50\% \\
    & \text{ w/o Prompting }             & 2.12 & 7.28 & 2.48\% \\
    & \textbf{ NRFormer }                & \textbf{1.99} & \textbf{6.24} & \textbf{2.31\%}\\

    \midrule
    \multirow{5}{*}{Japan-1D}
    & \text{ w/o Normalization }         & 2.18 & 9.05 & 2.14\% \\
    & \text{ w/o Temporal Attention }    & 2.25 & 9.68 & 2.44\% \\
    & \text{ w/o Spatial Attention }     & 2.26 & 9.33 & 2.52\% \\
    & \text{ w/o Prompting }             & 2.17 & 9.32 & 2.15\% \\
    & \textbf{ NRFormer }                & \textbf{2.01} & \textbf{8.76} & \textbf{2.13\%}\\
    \bottomrule
  \end{tabular}
  \vspace{-0.5cm}
\end{table}

\subsubsection{Implementation Details}
\label{sec:Implementation Details}
The dataset was partitioned into three distinct subsets: 60\% for training, 20\% for validation, and 20\% for testing. Our experimens are conducted on two different server configurations: Linux Centos with four RTX 3090 GPUs and Linux Ubuntu with two A800 GPUs.
In terms of implementation, our model is implemented using PyTorch, with the Adam optimizer chosen for optimization. We set the learning rate and the weight decay to 0.001 and 0.0001, respectively. The batch size is fixed to 32. The number of temporal and spatial attention layers are fixed to 3, and the head of the multi-head attention module is set to 4.

\subsection{Overall Performance}
Table~\ref{tab:overall results} presents the overall performance of NRFormer and all the baseline models on two real-world datasets with respect to MAE, RMSE, and MAPE. 
Overall, NRFormer consistently outperforms other baseline models on both datasets, demonstrating the effectiveness of NRFormer. 
Specifically, our model achieves improvements of 8.02\%, 10.95\%, 13.95\%, and 13.1\% beyond the LightCTS in terms of the MAE metric on 6, 9, 12, and 24 steps on the Japan-4H dataset, respectively. Similarly, the improvement of the MAE on the Japan-1D dataset is 7.57\%, 7.77\%, 8.5\%, and 6.51\%, respectively.
Moreover, we can make the following observations: (1) All traditional methods (\ie HA, LR, and XGBoost) perform worse than deep learning models, as they fail to capture spatio-temporal patterns. 
(2) STGNN-based approaches (\ie DCRNN, Graph WaveNet) outperform HA, LR, and XGBoost by a large margin, indicating the superiority of modeling intricate spatio-temporal dependencies. 
(3) STID performs worse than DCRNN and Graph WaveNet on Japan-1D dataset. The possible reason is that Japan-1D has fewer training samples than Japan-4H, which is insufficient to learn discriminative spatial and temporal identity embeddings.
(4) LightCTS leverages a tailored designed automatic spatio-temporal feature extraction module and exhibits strong learning capabilities, achieving the best performance compared with other baseline models.

Moreover, compared to sudden change experiments, NRFormer shows substantial improvements across all datasets. 
Specifically, compared to Koopa in sudden change forecasting, NRFormer achieves an (11.25\%, 14.32\%, 7.23\%) and (16.26\%, 9.85\%, 11.6\%) improvement on the Japan-4H and Japan-1D datasets, respectively, under the MAE, RMSE, and MAPE metrics.
The experimental results show that our model, by addressing spatiotemporal distribution imbalances and incorporating heterogeneous contextual factors, adapts more quickly to the non-stationary nature of radiation, providing more accurate predictions for sudden change samples.

\subsection{Ablation Study}
To verify the effectiveness of each component, we conduct ablation studies on four variants of the proposed NRFormer, including (1) \emph{w/o normalization} removes non-stationary normalization and denormalization strategy, (2) \emph{w/o temporal attention} replaces temporal attention layer by the linear layer, (3) \emph{w/o spatial attention} removes the imbalance-aware spatial attention module, and (4) \emph{w/o prompting} removes radiation propagation prompting module.
The results of all variants are shown in Table~\ref{tab:ablation_studies}.

As can be seen, NRFormer consistently outperforms all the variants, demonstrating the effectiveness of each proposed module. Specifically, the model performance degrades after removing the non-stationary normalization. This implies that incorporating instance-level normalization can attenuate the non-stationarity of radiation series and further enhance radiation forecasting accuracy. 
We observe that \emph{w/o temporal attention} and \emph{w/o spatial attention} perform worse than NRFormer, which indicates extracting the complex spatio-temporal features in radiation series is critical for forecasting. 
Furthermore, we also observe that the contextual prompt has a significant impact on model performance, indicating the advance of modeling radiation propagation with external factors.

We also examine the utility of the contextual features we provide to the prompting module. noPrompt means that we remove all the contextual features from our framework. 
Others (\eg Wind Speed) indicate using only one contextual feature.
The results on MAE and RMSE are presented in Figure~\ref{fig:prompt_sensitivity}. In the absence of any contextual features, we separately add one feature at each time to assess its contribution. Here \emph{noPrompt} indicates the performance of NRFormer without utilizing any additional contextual features.
We can observe that incorporating each feature consistently improves the model performance beyond the \emph{noPrompt}. In summary, the contextual features have a significant impact on nuclear radiation, and taking them into account is beneficial for forecasting.

\begin{figure}[t]
  \centering
  \includegraphics[width=1\linewidth]{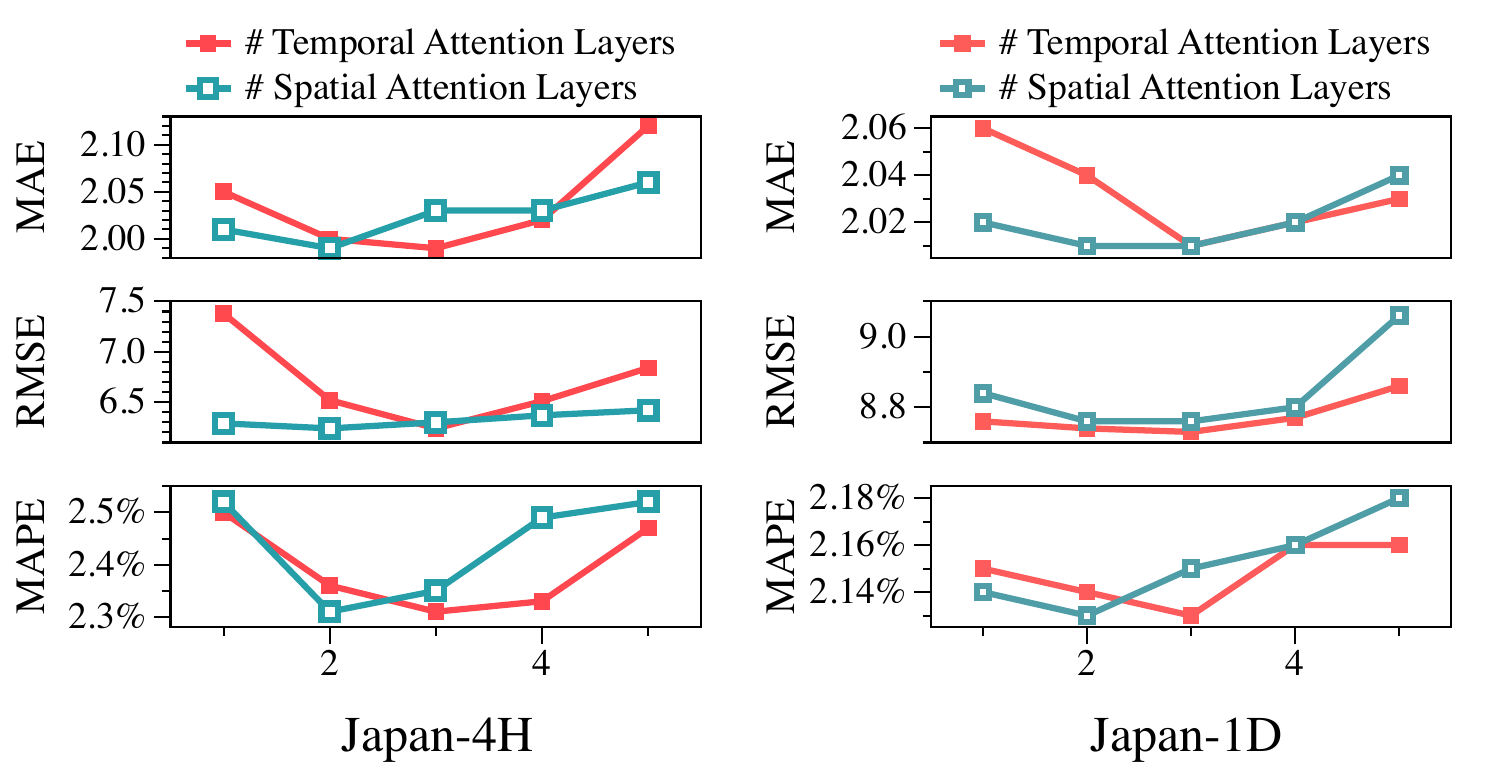}
  \vspace{-0.7cm}
  \caption{Parameter sensitivity of different temporal and spatial attention layers.}
  \label{fig:parameter_sensitivity}
  \Description{}
  \vspace{-0.5cm}
\end{figure}

\subsection{Parameter Sensitivity Analysis}
\label{parameter sensitivity analysis}
Finally, we study the impacts of the hyper-parameters on the performance of radiation forecasting. 
We evaluate the impact of the number of temporal attention layers $L_t$ and the number of spatial attention layers $L_s$. The results on MAE, RMSE, and MAPE are reported in Figure~\ref{fig:parameter_sensitivity}. 
First, we vary $L_t$ from 1 to 5. As can be seen in Figure~\ref{fig:parameter_sensitivity}, the performance shows a fluctuation when increasing $L_t$ from 1 to 3, and drops slightly by further increasing $L_t$ from 3 to 5 on both Japan-4H and Japan-1D datasets. Overall, we achieve the best performance when setting $L_t$ = 3. 
The possible reason is that a small $L_t$ is insufficient to capture temporal correlation information, whereas too large $L_t$ may introduce redundant and noisy information for our task, leading to performance degradation.
We vary $L_s$ from 1 to 5 and observe that our model achieves optimal performance when $L_s$ is set to 2. Increasing or decreasing $L_s$ beyond this point results in a decline in performance. This is primarily because a lower $L_s$ fails to provide adequate spatio-temporal correlation of radiation propagation. 
Additionally, we noticed a performance drop when more spatial attention layers were added. 
This phenomenon might be caused by the fact that in an imbalanced spatial distribution scenario, using an excessive number of $L_s$ can lead to overfitting, particularly for nodes with highly dense connections. 

\begin{figure}[t]
  \centering
  \includegraphics[width=1\linewidth]{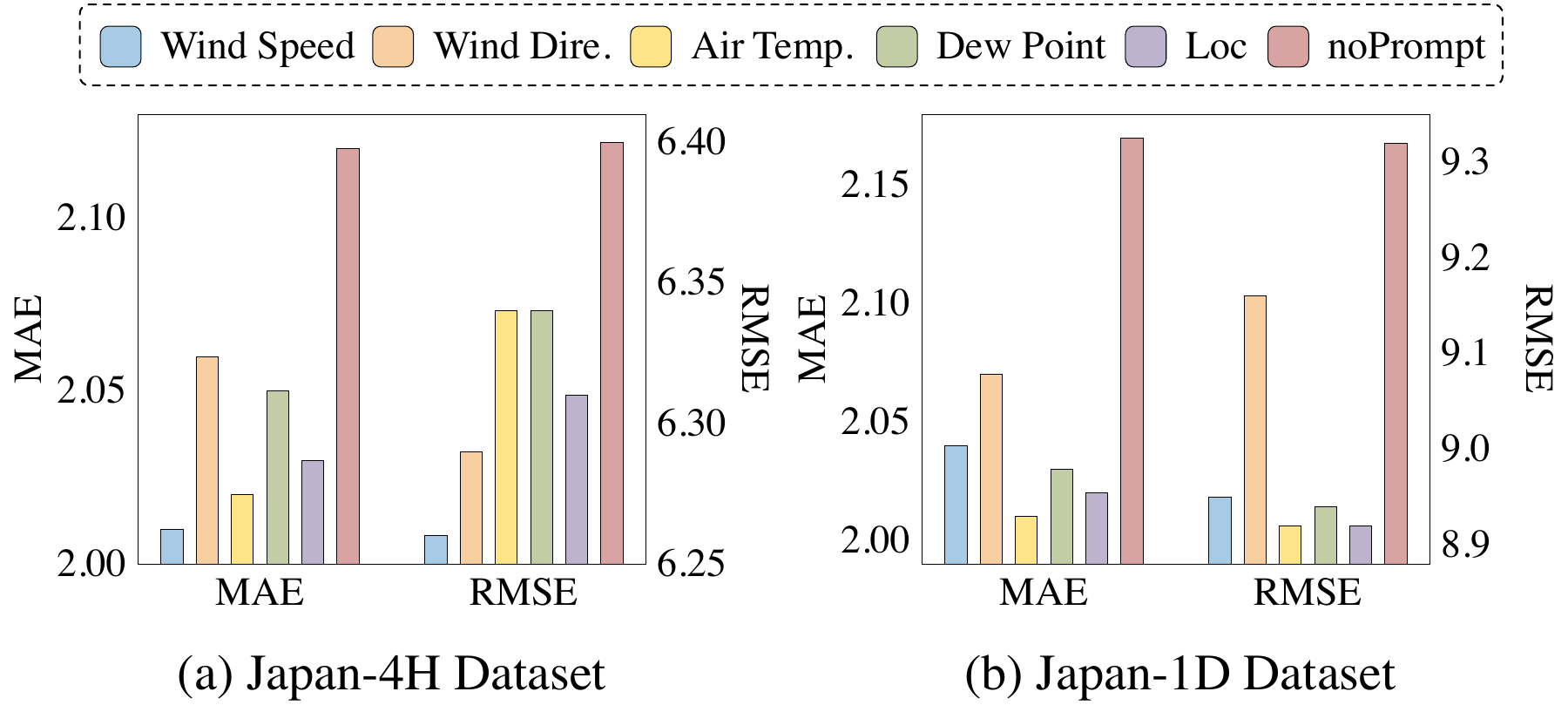}
  \vspace{-0.6cm}
  \caption{Ablation study of each contextual feature. }
  \label{fig:prompt_sensitivity}
  \Description{}
  \vspace{-0.6cm}
\end{figure}

\subsection{Deployed System Evaluation}
\begin{figure}[t]
  \centering
  \includegraphics[width=1\linewidth]{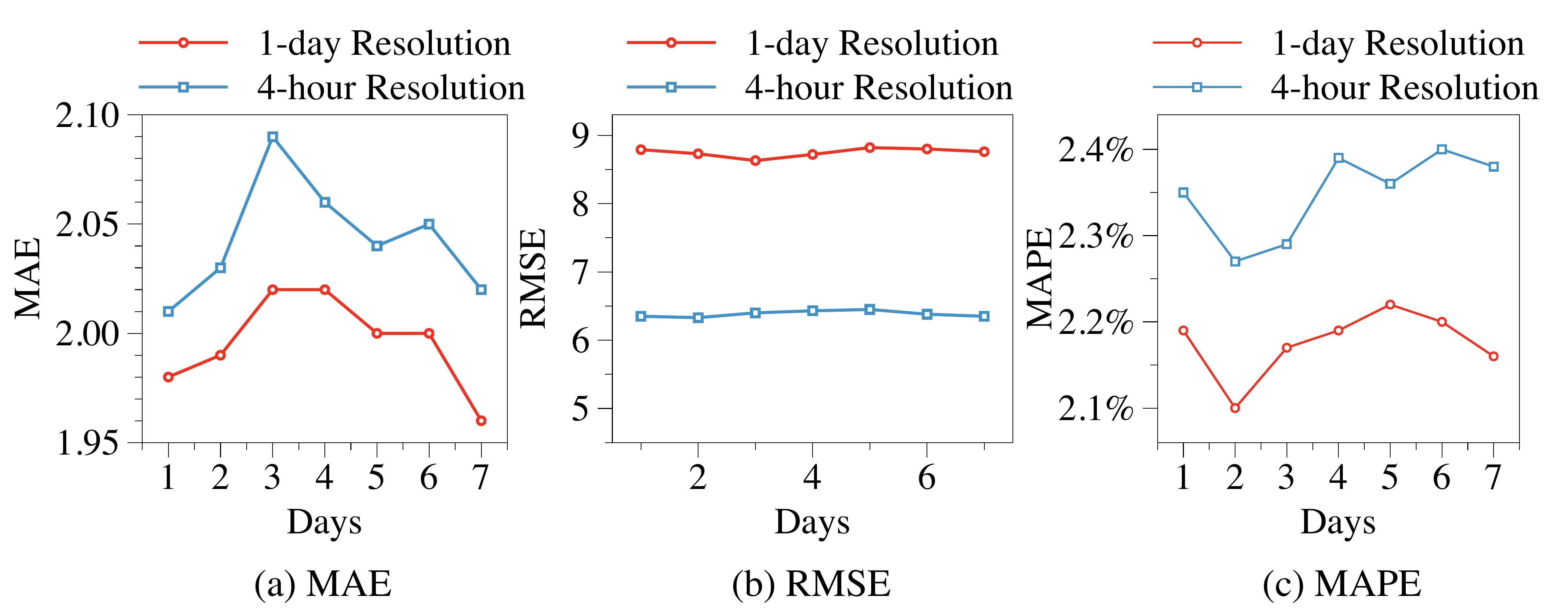}
  \vspace{-0.6cm}
  \caption{Performance evaluation of deployed radiation forecasting system using recent data from January 25-31, 2025.}
  \label{fig:system_evaluation}
  \Description{}
  \vspace{-0.6cm}
\end{figure}

To evaluate the real-world deployment performance of the NRFormer framework, we conducted a rigorous temporal generalization analysis using the most recent radiation monitoring data. As shown in Figure \ref{fig:system_evaluation}, the system maintained robust predictive capabilities during the critical evaluation window of January 25–31, 2025, achieving high accuracy in seven-day rolling averages despite significant temporal distributional shifts between the training period (March 17, 2021 – January 1, 2024) and the testing phase. This performance demonstrates the system’s strong stability and robustness, thereby highlighting its practical utility for both individual users and government agencies by empowering them to make timely, data-driven decisions regarding emergency response and public safety.

\section{Related Works}
\label{sec:Related Works}
This section emphasizes related works in two aspects: transformer and spatio-temporal forecasting.

\textbf{Transformer.}
The advent of the Transformer architecture~\cite{vaswani2017attention} has been making a transformative impact across various fields, particularly in time series forecasting~\cite{zhou2021informer,cao2023inparformer,chowdhury2022tarnet,gorbett2023sparse, wutimesnet, wang2023micn, jia2024witran} and graph mining~\cite{zhang2021mugrep, ZhuWS0Z23, fan2021heterogeneous, li2022kpgt, wu2022nodeformer, chen2021z, wang2023networked}. The transformer depends on the self-attention mechanism to capture data correlations and is capable of modeling complicated and long-range dependencies. 
Due to the powerful capability, recently considerable endeavors have been made to adapt Transformer to time series and spatio-temporal data~\cite{zha2021shifted,feng2022adaptive,liang2023airformer,zhang2022crossformer,liu2023spatio, li2023revisiting}.
For instance, by introducing a ProbSparse self-attention mechanism, Informer~\cite{zhou2021informer} enhances the Transformer framework for time series forecasting with high efficiency and scalability. 
STAEformer~\cite{liu2023spatio} revolutionizes time series forecasting by adeptly capturing spatio-temporal dynamics, setting new benchmarks in accuracy on real-world datasets with its unique embedding technique.
As another example, Autoformer\cite{wu2021autoformer} further pushes the boundaries of time series forecasting by integrating decomposition and auto-correlation mechanism into Transformer. In this paper, we adapt the Transformer model with tailored designs to address the unique challenges in nuclear radiation forecasting.

\textbf{Spatial-temporal forecasting.}
Spatio-temporal forecasting aims to predict the future states of spatio-temporal systems, such as traffic~\cite{feng2022adaptive,liu2022practical,ji2023spatio,jiang2023spatio} and atmospheric~\cite{han2021joint,liang2023airformer} systems, based on historical observations. 
Nuclear radiation forecasting can be naturally modeled as a spatio-temporal forecasting problem.
In recent years, Spatio-Temporal Graph Neural Networks (STGNNs)~\cite{jin2023spatio,han2021dynamic,shao2022pre} has emerged as the most prevalent approaches in this field due to their strong capability in capturing intricate spatio-temporal dynamics. 
To name a few, DCRNN~\cite{li2018diffusion} models the diffusion process of traffic flows by leveraging graph-based diffusion convolution coupled with recurrent neural networks (RNNs).
Graph Wavenet~\cite{wu2019graph} learns an adaptive graph structure to capture the latent spatial relationships by decomposing the adjacency matrix into two learnable embedding matrices. 
In terms of radiation forecasting, existing studies rely on either numerical simulation~\cite{kim2005use} or statistical methods~\cite{khanna1996nuclear,egarievwe2016analysis,sun2022spatial} to predict future radiation levels. However, these methods struggle with modeling complex spatio-temporal dependencies, leading to degraded performance in large-scale prediction scenarios.

\section{Conclusion}
\label{sec:Conclusion}
In this paper, we introduce NRFormer, a spatio-temporal graph Transformer for national-wide nuclear radiation forecasting. Specifically, we first devise a non-stationary temporal attention module to capture the erratic and non-stationary characteristics in radiation time series. Subsequently, we propose an imbalance-aware spatial attention module to address the imbalanced spatial distribution issue by adaptively re-weighting the influence of stations from both macro-scale and proximity-constrained perspectives. Furthermore, a radiation propagation prompting module is developed to guide the predictive modeling process.
Extensive experiments on real-world nuclear radiation datasets demonstrate the superiority of NRFormer against seven baselines. In the future, we plan to deploy NRFormer to more countries so as to empower various decision-making tasks such as emergency response planning, thereby protecting environmental safety and public health.

\begin{acks}
This work was supported by the National Key R\&D Program of China (Grant No.2023YFF0725004), National Natural Science Foundation of China (Grant No.92370204), the Guangzhou Basic and Applied Basic Research Program under Grant No. 2024A04J3279, Education Bureau of Guangzhou Municipality.
\end{acks}

\bibliographystyle{ACM-Reference-Format}
\bibliography{sample-base}

\clearpage
\appendix

\begin{figure*}[t]
  \centering
  \includegraphics[width=1\linewidth]{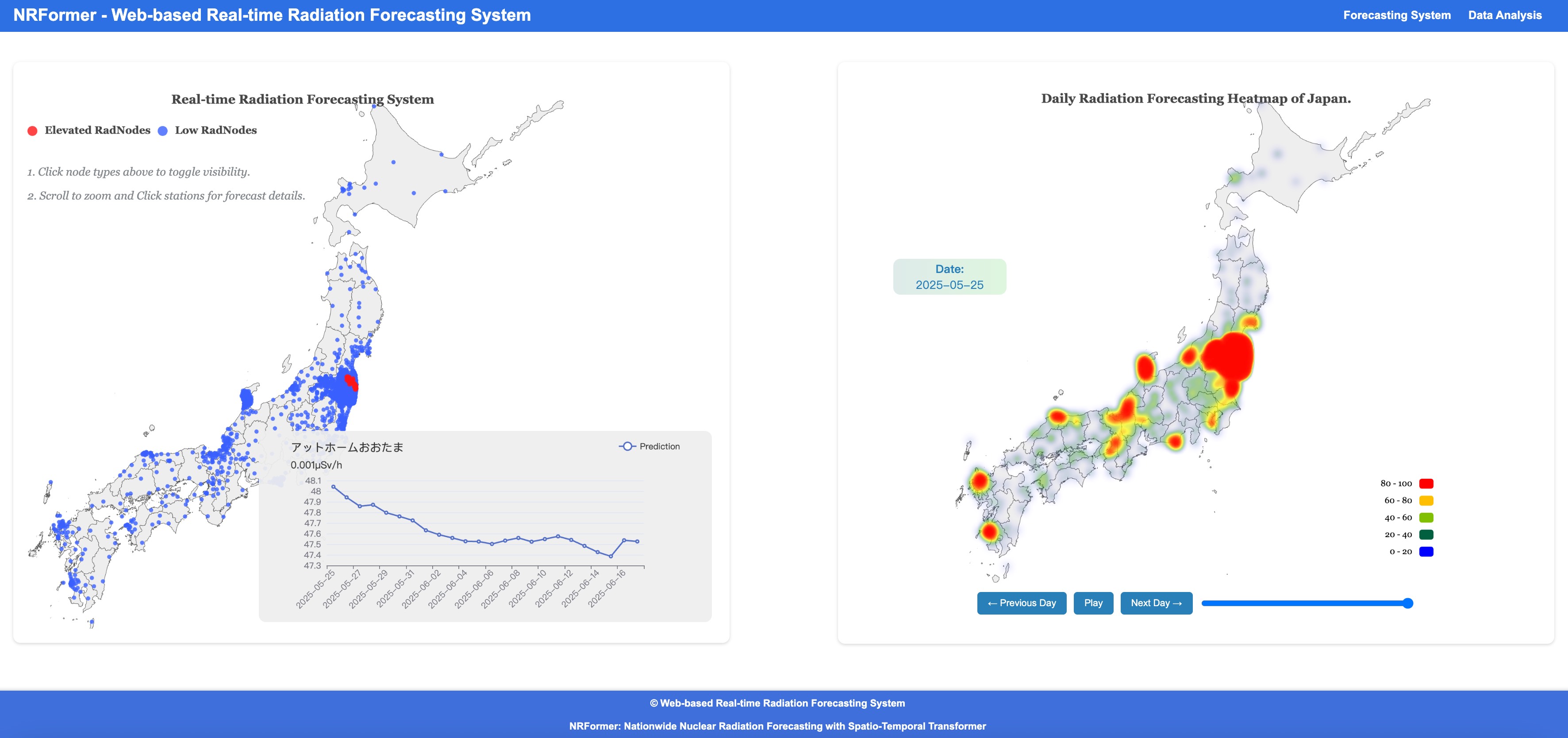}
  \caption{The deployed web-based radiation monitoring and forecasting system. Left panel: Forecasts for individual radiation monitoring stations, with interactive stations enabling users to access detailed predictions for the coming days. Right panel: Radiation dispersion dynamics across Japan, highlighting spatiotemporal propagation patterns over the next several days.}
  \label{fig:appendix_forecasting}
  \Description{}
  \vspace{1cm}
\end{figure*}

\begin{figure*}[t]
  \centering
  \includegraphics[width=1\linewidth]{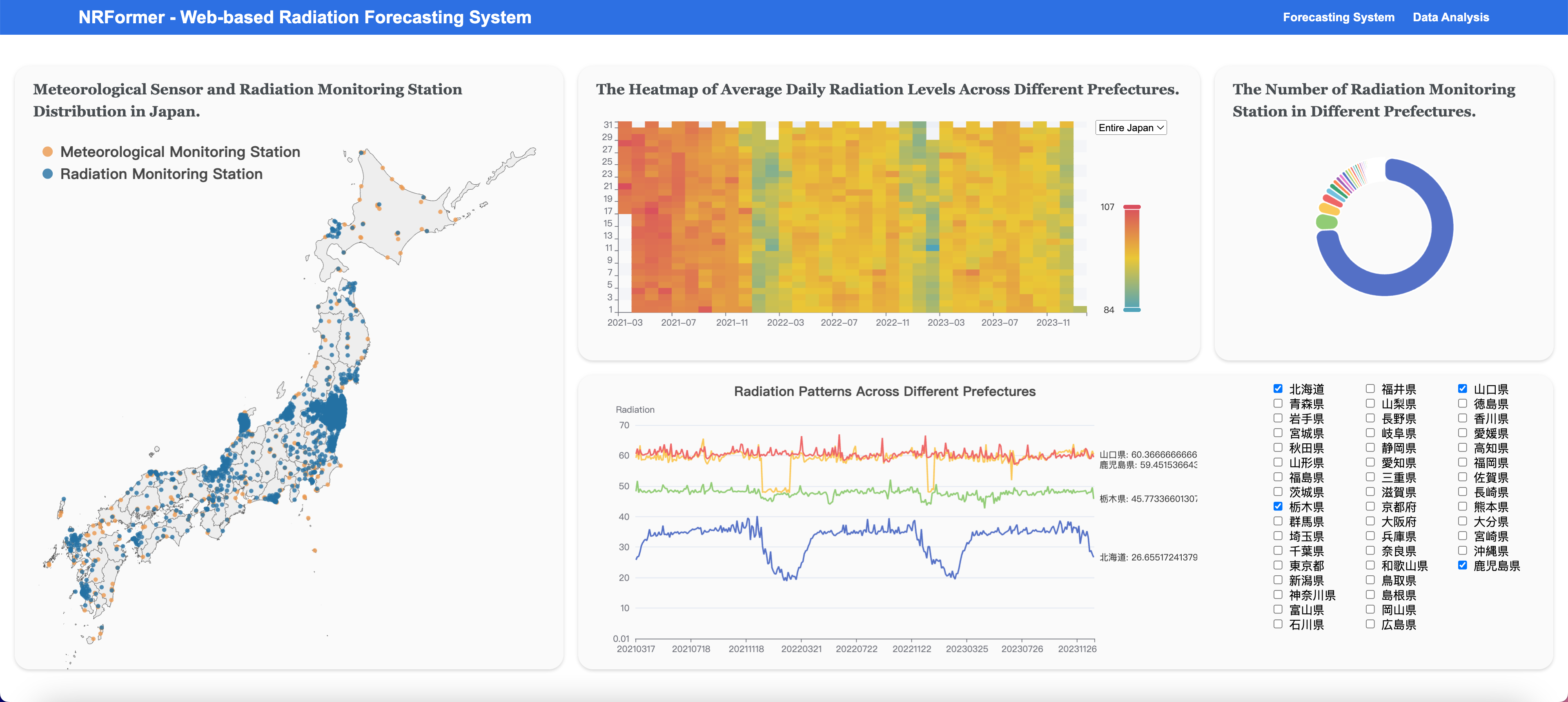}
  \caption{The data visualization of nuclear radiation in Japan: the left panel shows the distribution of radiation and meteorological monitoring stations; the top-center chart illustrates the average daily radiation levels across different prefectures; the top-right chart depicts the number of radiation monitoring stations in each prefecture; and the bottom-right chart presents temporal radiation patterns observed across various prefectures.}
  \label{fig:appendix_data_vis}
  \Description{}
\end{figure*}

\end{document}